# Visible Light-Based Human Visual System Conceptual Model

Lee Prangnell
Department of Computer Science, University of Warwick, England, UK
lee.prangnell@warwick.ac.uk

**Abstract** — There exists a widely accepted set of assertions in the digital image and video coding literature, which are as follows: the Human Visual System (HVS) is more sensitive to luminance — often confused with brightness — than photon energies (often confused with chromaticity and chrominance). Passages similar to the following occur with high frequency in the peer reviewed literature and academic text books: "the HVS is much more sensitive to brightness than colour" and/or "the HVS is much more sensitive to luma than chroma".

In this discussion paper, a Visible Light-Based Human Visual System (VL-HVS) conceptual model is discussed. The objectives of VL-HVS are as follows: 1. To provide a deeper theoretical reflection of the fundamental relationship between visible light, the manifestation of colour perception derived from visible light and the physiology of the perception of colour. That is, in terms of the physics of visible light, photobiology and the human subjective interpretation of visible light, it is appropriate to provide comprehensive background information in relation to the natural interactions between visible light, the retinal photoreceptors and the subsequent cortical processing of such. 2. To provide a more wholesome account with respect to colour information in digital image and video processing applications. 3. To recontextualise colour data in the RGB and YCbCr colour spaces, such that novel techniques in digital image and video processing — including quantisation and artifact reduction techniques — may be developed based on both luma and chroma information (not luma data only).

## 1.0 Physics of Electromagnetic Radiation (Visible Light Range)

In terms of visible light emissions from natural and synthetic light sources, luminance and photon energies are observable phenomena; they interact simultaneously in the retinal photoreceptor system, after which the visible light information is processed in the visual cortex in the brain. This results in an intertwined perception of brightness and colourfulness [1, 2]. Note that both brightness and colourfulness exist only in the perceptual domain and are, thus, not empirically observable in nature. Among other things, human perception of brightness depends on both luminance and photon energy; likewise, human perception of colourfulness depends on both luminance and photon energy.

All forms of electromagnetic radiation in the electromagnetic spectrum, from radio waves through to gamma rays, are different manifestations of light. According to Albert Einstein's theory of special relativity, all forms of electromagnetic radiation — including infrared light, visible light and ultraviolet light — travel at the same velocity (in a vacuum), which is presently established as 299,792,458 metres per second (m/s) [3, 4]. This figure is typically approximated to $3\times10^8$ m/s when performing calculations. According to Einstein's theory of special relativity, the speed of light cannot be exceeded.

**Table 1**. Photon wavelength, frequency and energy of visible light in the electromagnetic spectrum. This range of wavelengths, frequencies and energies manifests as a range of colours — and the different subjective properties of colour — in the visual systems of humans, African monkeys, apes and chimpanzees.

| Perceptual Manifestation | Wavelength | Frequency | Energy |
|---|---|---|---|
| Violet | 380–450 nm | 668–789 THz | 2.75–3.26 eV |
| Blue | 450–495 nm | 606–668 THz | 2.50–2.75 eV |
| Green | 495–570 nm | 526–606 THz | 2.17–2.50 eV |
| Yellow | 570–590 nm | 508–526 THz | 2.10–2.17 eV |
| Orange | 590–620 nm | 484–508 THz | 2.00–2.10 eV |
| Red | 620–750 nm | 400–484 THz | 1.65–2.00 eV |

Visible light is a portion of light in the electromagnetic spectrum — within a narrow range of specific wavelengths — that is visible to human beings via the HVS. The established wavelength range of visible light is approximately 380 to 750 nanometres (nm), which equates to a frequency range of 668 to 484 terahertz (THz), respectively [5, 6, 7]. At the quantum level, the fundamental particle of light, including visible light, is known as the photon. The photon has been shown to behave simultaneously as a particle and a wave; in quantum mechanics research, this is referred to as the counterintuitive wave-particle duality phenomenon. Moreover, the photon is an elementary particle in the boson category. The energy propagated by an electromagnetic wave, for all forms of electromagnetic radiation including visible light, is continuously distributed in the form of photons. The photon energy of visible light ranges from 2 to 2.75 electron volts (eV) [5, 6, 7]; the energy of a photon is inversely proportional to the wavelength of the electromagnetic wave (see Table 1).

Colour vision in humans can be considered as the combined visual perception of the different luminances and photon energies emitted from either natural or synthetic visible light sources. Every aspect of colour that humans visually perceive is ultimately contingent upon the natural processes of visible light and the subsequent biological processing of such. In other words and in simple terms, colour is the subjective interpretation of electromagnetic radiation in the spectrum of visible light.

As previously mentioned, the photon acts as both a wave and a particle. Photon energy $E$ can be measured in either joules (J) or eV [5, 6, 7], and is quantified in (1):

$$E = \frac{h \cdot c}{\lambda} \qquad (1)$$

where $h$ and $c$ are universally accepted constants in the field of physics. Planck's constant $h = 6.626 \times 10^{-18}$ joule seconds (J·s), which computes the quantum of action. Constant $c = 3 \times 10^{8}$ metres per second (m/s), which quantifies an approximation of the speed of light in a vacuum. Lambda $\lambda$ in this context corresponds to the wavelength of a photon in metres. Note that $1\ \text{J} = 6.242 \times 10^{18}$ eV.

Utilising the formula in (1), we can quantify the photon energy of visible light electromagnetic radiation that humans perceive as red (red light), $E_{\text{red}}$, assuming the wavelength $\lambda = 700$ nm, in (2).

$$E_{\text{red}} = \frac{\left(6 \times 10^{-34}\,\text{J} \cdot \text{s}\right) \times \left(3 \times 10^{8}\,\text{m/s}\right)}{700 \times 10^{-9}\,\text{m}} = 2.8 \times 10^{-19}\,\text{J} = 1.8\text{eV} \tag{2}$$

As shown in (2), assuming the wavelength of 700 nm, the energy of a photon of red light is approximately 1.8eV (see Table 1). The number of photons $N$ emitted per second by a visible light source is contingent upon the energy of the visible light source $P$, measured in joules (i.e., J/s×1s), and the energy of the photon [5, 6, 7], which is given by (3).

$$N = \frac{P}{E} \tag{3}$$

Photon flux $\Phi_P$ is a term used to describe the number of photons emitted per unit area per unit time (i.e., $m^2/s$). The photon intensity is the photon flux per unit solid angle. The photon flux is computed in (4).

$$\Phi_{\text{P}} = \frac{N}{\text{m}^2/\text{s}} \tag{4}$$

The luminous intensity $I_v$ of a visible light source is quantified as the wavelength weighted power emitted from the source in a direction per solid angle, which is measured in candela (cd). Luminous intensity corresponds to the intensity of visible light and is, thus, computed based on the wavelength $\lambda$ of a photon, as shown in (5):

$$I_{\text{V}} = C \cdot V\left(\lambda\right) \cdot I_{\text{e}} \tag{5}$$

where $V(\lambda)$ is the luminosity function, also known as the luminous efficiency function, which is standardised by The International Commission on Illumination. $V(\lambda)$ quantifies the average spectral sensitivity of luminance in the human eye [8]. $C$ corresponds to the constant value of 683 lumens per watt (lm/W), $I_e$ refers to the radiant intensity, which is measured in watts per steradian (W/sr). Note that $V(\lambda)$ is utilised to define the luminous flux, $\Phi_L$, which is measured in lm and is computed in (6):

$$\Phi_{\text{L}} = C \cdot \int_{0}^{\infty} V\left(\lambda\right) \cdot \Phi_{\text{e},\lambda}\left(\lambda\right) \cdot d\left(\lambda\right) \tag{6}$$

where $\Phi_{e,\lambda}$ is the spectral radiant flux, measured in watts per nanometre (W/nm). Recall that variable $\lambda$ corresponds to the wavelength of the photon.

Luminance is a measurement of luminous intensity travelling in a given direction; it is measured in candela per square metre ($cd/m^2$). From the perspective of light emitting from a light source, such as a TV, a Visual Display Unit (VDU) or an incandescent light bulb, luminance computes the light emitted from the light source, which is then distributed within a solid angle [5, 6, 7]. The luminance within a ray of light, denoted as $L_v$, is computed in (7):

$$L_V = n^2 \frac{d\Phi_L}{G} \tag{7}$$

where $n$ corresponds to the index of refraction of an object, $d\Phi_L$ refers to the luminous flux carried by the beam of the light source and $G$ denotes étendue of a narrow beam containing the ray.

The inverse square law must be taken into account because it describes the distribution and the intensity of visible light over arbitrary macroscopic distances. For example, energy that is shown to be twice the distance from the visible light source is spread over four times the area from the source, which equates to the distributed visible light being one fourth the intensity of the visible light in the source. This can be quantified by computing the illuminance $M$ — the amount of luminous flux per unit area — which is quantified in (8):

$$M = \frac{T}{r^2} \tag{8}$$

where $T$ is the power unit per solid angle, also known as pointance. $T$ is computed in (9):

$$T = \frac{S}{A} \tag{9}$$

where $S$ is the strength of the visible light source, which can be measured in terms of power, for example, and where $A = 4\pi r^2$, which corresponds to the sphere area of the visible light source.

**2.0 Human Visual System**

From the perspective of the Darwinian paradigm of evolutionary biology, the HVS is the product of billions of years of evolution by natural selection [9]. It is a hugely complex phenomenon, for which scientific models endeavour to rigorously conceptualise the interaction between visible light, the eye and the corresponding perceptual processing of visible light — and the colours derived from such — in the human brain. In psychophysical research, there exists extensive evidence which highlights the fact that colour vision is facilitated by the interaction of photons with the retinal photoreceptor systems, whereby such photons are biologically converted into electrical signals in the retina (known as visual phototransduction) [10, 11, 12].

With a focused concentration on retinal photoreceptors, which are shared by all species in the taxonomic order of primate (including humans), the key photoreceptors are as follows: rods and cones. The retinal photoreceptor system is dominated by rods (120,000,000 units) compared with cones (6,400,000 units). Rods are specialised for low visible light conditions [10, 11, 12]. When subjected to higher intensities of visible light the transmitter release stops because the rod's response to the visible light is much slower than the cone's response. Cones are the retinal photoreceptors that facilitate colour vision and colour perception. They allow for fast, high acuity and colour vision because they are able to adapt to a vast variety of visible light intensities [10, 11, 12].

In terms of the population of cones, empirical experiments have revealed that 64% are sensitive to photons that are perceived as red, 32% green and 4% blue (trichromatic colour vision). There are three classifications of retinal cone: Long (L), Medium (M) and Short (S), each of which contains the transmembrane protein opsin and the molecule chromophore, which are the constituents of photosensitive visual pigments. These pigments are especially sensitive to photons within the following approximate wavelength ranges: 650 nm (L), 510 nm (M) and 475 nm (S), which humans interpret as red, green and blue, respectively [10, 11, 12]. This aspect of the HVS catalysed the emergence of the RGB colour model. The human visual perception of colour and the different characteristics of such are, among other things, contingent upon the level of excitation of the different cones. Furthermore, in general terms, the visual cortex system in the brain is responsible for differentiating the signal response received from the L, M, S cones and, thus, results in the discernment of a vast range of signals, which are consequently perceived in the form of a wide range of colours (wide colour gamut). The science behind this process spawned the creation of the CIE 1931 colour space chromaticity diagram, developed by The International Commission on Illumination (CIE) [13].

## 3.0 RGB and YCbCr Colour Spaces

The RGB colour model is an additive tristimulus colour model that is ubiquitous in a plethora of computer science and consumer electronics applications. In basic terms, it amalgamates colour from the following primary colours: Red (R), Green (G) and Blue (B), which results in a range of colours depending on the corresponding pixel intensity, colour gamut and bit depth permitted [14, 15, 16]. The scientific basis for the creation of the RGB colour model, and the subsequent RGB colour space utilised in digital imaging, is the Young-Helmholtz theory of trichromatic colour vision, which is congruent to the aforementioned L, M, and S cone sensitivity to the wavelengths of photons that humans perceive as red, green and blue, respectively. In terms of pixels on display devices, including the Cathode Ray Tube (CRT), the Liquid Crystal Display (LCD) and the Light Emitting Diode (LED), each pixel in CRT, LCD and LED VDUs and TVs comprise RGB light sources. Therefore, the physics of visible light and the biology of colour vision, explained in sections 1.1 and 1.2, respectively, will always apply. That is, the interaction of the HVS with luminance and photon energy, emitted from the VDU or TV (light source), is the physical process by which the electromagnetic radiation — in the visible light spectrum — is visually perceived as brightness and colourfulness in the human observer.

In basic terms, RGB-based colour can be represented by either normalised arithmetic, percentage or base-10 integer representations of binary numbers. Expressed as a triplet, the binary representations of $R$, $G$ and $B$ data are dependent on the bit-depth of each channel. There are $2^b-1$ pixel intensities in each colour channel, where $b$ corresponds to the bit-depth. For a bit-depth of 8 bits per pixel per channel (24 bits per pixel), the integer values ranges are as follows: $R \in [0,255]$, $G \in [0,255]$ and $B \in [0,255]$. In this example, $R$=0, $G$=0 and $B$=0 represents absolute black (low energy). Conversely, $R$=255, $G$=255 and $B$=255 represents absolute white (high energy). For image or video data with higher bit-depths (e.g., 16-bit data), this equates to a vastly greater number of colours in each pixel. For 16-bit image or video data, the integer value ranges are as follows: $R \in [0,65535]$, $G \in [0, 65535]$ and $B \in [0,65535]$, where $R$=65535, $G$=65535 and $B$=65535 represents absolute white [14, 15, 16].

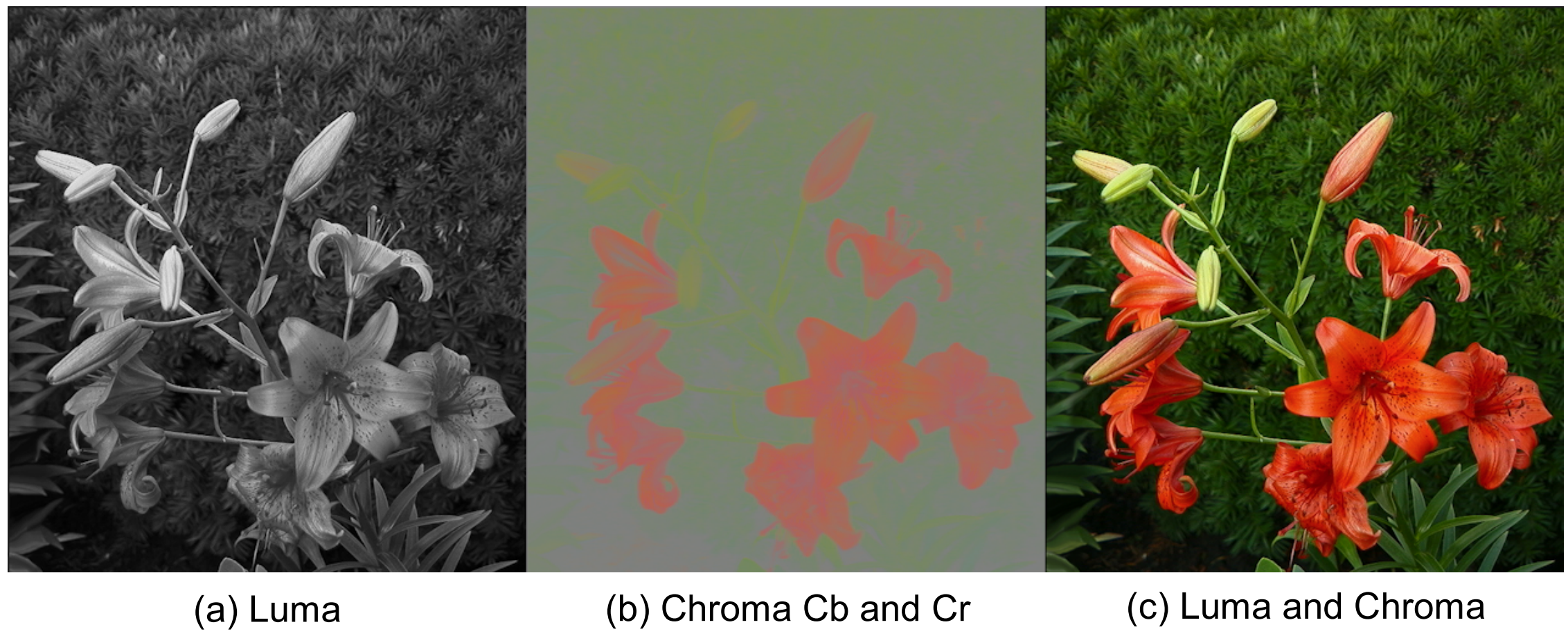

**Figure 1**. A subsampled, uncompressed 8-bit YCbCr (4:2:0) colour digital image showing (a) the Y (luma) channel, (b) the Cb and Cr (chroma) channels combined and (c) the image in which the constituent luma and chroma data is combined. Note the conspicuousness of finer detail of the image — and also the monochromacy — in the luma channel.

The YCbCr colour space is a colour transformation from a given RGB colour space comprising luma and chroma channels. It is often confused with the YUV colour space; they are, however, very closely related. YCbCr is designed for digital imaging and video, whereas YUV is designed for analog video. YCbCr is, thus, a scaled version of YUV.

Luma and chroma are colloquial terms typically used only in the image and video processing community. Luma (Y) refers to an achromatic colour channel that is derived via an approximation of gamma-corrected luminance. The human perception of the brightness of the colour in the luma channel is conceptualised as relative luminance, where the values are normalised to 1 or 100. On a percentage scale, 0% represents absolute black and 100% represents absolute white. Moreover, the luma channel contains the vast majority of the finer detail in an image (see Figure 1). Chroma is short for chrominance (not chromaticity). Chrominance (Cb and Cr) refers to the colour difference channels, which are as follows: blue difference (Cb) and red difference (Cr) with reference to the luma (Y) channel; Cb and Cr collectively correspond to the perceptual colourfulness of the colour in an image [14, 15, 16].

As regards the ubiquitous assumptions concerning the sensitivity of the HVS to luma and chroma data, it is certainly true that the HVS is vastly more sensitive to the low pass filtering (such as Gaussian blur) of high frequency detail in luma sample data. This is because data in the luma channel contains almost all of the finer details in a picture (see Fig. 1). As such, chroma samples can be spatially downsampled because they can tolerate a much greater level of low pass filtering. The high HVS sensitivity to certain gradations of luma samples, in YCbCr data, should not be confused with HVS sensitivity to luminance and photon energy.

In a mathematical sense, $Y$, in YCbCr, which is referred to as luma, corresponds to the weighted sum of RGB values (not gamma corrected). The gamma corrected version is denoted as $Y'$. Concentrating on the gamma corrected version, $Y'$, from ITU-R BT.2020 [17], is computed in (10):

$$Y' = (0.2627 \cdot R') + (0.6780 \cdot G') + (0.0593 \cdot B') \tag{10}$$

The parameter values applied to the gamma corrected RGB data ($R'$, $G'$ and $B'$), as shown in (10), are weights determined by the luminosity function as described in (5). It has been shown, in empirical testing, that humans are perceptually more sensitive to green in terms of brightness perception, in which case $G'$ is assigned the largest weight. Conversely, $B'$ is assigned the lowest weight.

$Cb'$ and $Cr'$ are colour difference channels with reference to the $Y'$ colour channel. The gamma corrected versions of chroma Cb and Cr, from ITU-R BT.2020 [17], are computed in (11) and (12), respectively:

$$Cb' = \frac{B - Y'}{1.8814} \tag{11}$$

$$Cr' = \frac{R - Y'}{1.4746} \tag{12}$$

With relevance to the compression of video and still image data, the ubiquitous conception that the HVS is more sensitive to luma data than chroma data typically results in a widely held assumption that compression artifacts (assuming equal levels of compression) are always more conspicuous in reconstructed luma data. While this may be the case in certain instances, it may not be the case in other instances. We have established that the HVS is incontestably more sensitive to the low pass filtering — e.g. Gaussian blur — of high frequency detail in luma data. However, this does not necessarily equate to the fact that the HVS is more sensitive to, for example, quantisation-induced compression artifacts in luma data. This is especially true in situations where artificial content and aggressive chroma subsampling is concerned; for example, the aggressive quantisation and spatial downsampling of YCbCr 4:4:4 vector graphics where the detail in the chroma data is of considerable importance.

## 4.0 Conclusions

With a focused concentration on contemporary image and video coding, the HVS is perceptually impervious to small gradations in blocks of samples in the Cb and Cr chroma channels, such as reasonable spatial downsampling and mild to moderate low pass filtering. However, this is not necessarily the case in situations where aggressive quantisation is applied. One good example of this is evident in the High Efficiency Video Coding standard [18]: for YCbCr 4:2:0 input video data, there is an intrinsic Quantisation Parameter (QP) offset mechanism in the HEVC reference software whereby chroma prediction residue is quantised at lower levels compared with the levels of quantisation applied to the corresponding luma prediction residue [19-23]. In this context, the largest chroma QP permitted is $QP$=39; conversely, the largest luma QP permitted is $QP$=51. Note that this chroma QP offset mechanism is not applicable to YCbCr 4:2:2 and 4:4:4 input video data [19-23]. Finally, the information provided in this discussion paper may catalyse the development of visually lossless compression contributions that consistently account for the data in all colour channels.